\documentclass[letterpaper, 10 pt, journal]{fmt/IEEEtran}
\IEEEoverridecommandlockouts                               

\usepackage{graphics} 
\usepackage{epsfig}   
\usepackage{mathptmx} 
\usepackage{times}    
\usepackage{amsmath}  
\usepackage{amssymb}  
\usepackage{mathtools}
\usepackage{verbatim}
\usepackage{cite}
\usepackage{arydshln}
\usepackage{subfigure}
\usepackage{booktabs}
\usepackage[ruled,linesnumbered]{algorithm2e}
\usepackage{tikz}
\usepackage{scalefnt}
\usetikzlibrary{positioning,shapes}
\SetKwComment{Comment}{}{}

\newtheorem{Def}{Definition}

\DeclareMathAlphabet\mathcal{OMS}{cmsy}{m}{n}
\SetMathAlphabet\mathcal{bold}{OMS}{cmsy}{b}{n}

\usepackage{longtable}
\usepackage{tabu}
\newsavebox\ltmcbox
\usepackage{booktabs}
\usepackage{tabularx}
\usepackage{multirow}
\usepackage{bm}
\usepackage{caption}

\usepackage{pgf,tikz,pgfplots}
\pgfplotsset{compat=1.15}
\usepackage{mathrsfs}
\usetikzlibrary{arrows}
\usetikzlibrary{decorations.markings}

\DeclareMathOperator{\acos}{acos}

\usepackage{wasysym}

\newcommand{\etal}{\textit{et al}. }
\newcommand{\plucker}{Pl$\mathrm{\ddot{u}}$cker }
\newcommand{\Log}{\textrm{Log}}

\makeatletter
\let\NAT@parse\undefined
\makeatother
\usepackage{hyperref}
\usepackage[square,numbers]{natbib}

\usepackage{amsmath,amssymb,graphicx}

\renewcommand{\parallel}{\mathrel{/\mkern-5mu/}}
\makeatletter
\newcommand{\notparallel}{%
  \mathrel{\mathpalette\not@parallel\relax}%
}
\newcommand{\not@parallel}[2]{%
  \ooalign{\reflectbox{$\m@th#1\smallsetminus$}\cr\hfil$\m@th#1\parallel$\cr}%
}
\makeatother

\title{\LARGE \bf
Calibration System and Algorithm Design for a Soft Hinged Micro Scanning Mirror with a Triaxial Hall Effect Sensor
}

\author{Di~Wang, Xiaoyu~Duan, Shu-Hao~Yeh, Jun~Zou, and Dezhen~Song
\thanks{D. Wang, S. Yeh and D. Song are with CSE Department, Texas A\&M University, College Station, TX 77843, USA. D. Song is also with Dept. of Robotics, Mohamed Bin Zayed University of Artificial Intelligence (MBZUAI), Abu Dhabi, UAE. Email: \texttt{dezhen.song@mbzuai.ac.ae}.}
\thanks{X. Duan and J. Zou are with ECE Department, Texas A\&M University, College Station, TX 77843, USA, Email: \texttt{junzou@tamu.edu}. }
\thanks{This work was supported in part by National Science Foundation under NRI-1925037 and Amazon Research Award 2020.}
}

\begin{document}

\maketitle
\thispagestyle{plain}
\pagestyle{plain}
\begin{abstract}
Micro scanning mirrors (MSM) extend the range and field of view of LiDARs, medical imaging devices, and laser projectors. However, a new class of soft-hinged MSMs contains out-of-plane translation in addition to the 2 degree-of-freedom rotations, which presents a cabliration challenge. We report a new calibration system and algorithm design to address the challenge. In the calibration system, a new low-cost calibration rig design employs a minimal 2-laser beam approach. The new new algorithm builds on the reflection principle and an optimization approach to precisely measure MSM poses. To establish the mapping between Hall sensor readings and MSM poses, we propose a self-synchronizing periodicity-based model fitting calibration approach. We achieve an MSM poses estimation accuracy of
$0.020^\circ$ with a standard deviation of $0.011^\circ$.   
\end{abstract}

\section{Introduction} \label{sec:intro}
Micro scanning mirrors (MSMs) are important component of active sensing, and they can extend the range and field of view (FoV) of LiDARs \cite{Wang'20}, medical imaging devices \cite{Pengwang'16}, and laser projectors \cite{Holmstrom'14}. Although existing research has explored techniques for calibrating MSMs with pure 2 degrees of freedom (DoF) in rotation, a full 3-DoF MSM motion model with additional translation pointing out of the mirror plane has not been well studied~\cite{schwarz'20, kim'17,li'19, yoo'21,Rembe'01-0}. In fact, the additional translation cannot be ignored in soft-hinged MSMs which are built upon hinges made of soft material instead of rigid revolute joints. 
Simultaneously measuring the 2-DoF rotation and 1-DoF translation of MSMs remains challenging because
1) the three types of motion's influence on reflection are coupled and cannot be recovered separately by observing a single reflected point, and
2) the small mirror surface area and large dynamic scanning range of MSMs make recovering 3-DoF motions from direct observations of the mirror plane impractical.

\begin{figure}[!htbp]
    \centering
    \subfigure[]{\includegraphics[width=2in]{image/MirrorModel.pdf}\label{fig:mirror}}\\
    \vspace{-.1in}
    \subfigure[]{\includegraphics[height=1.26in]{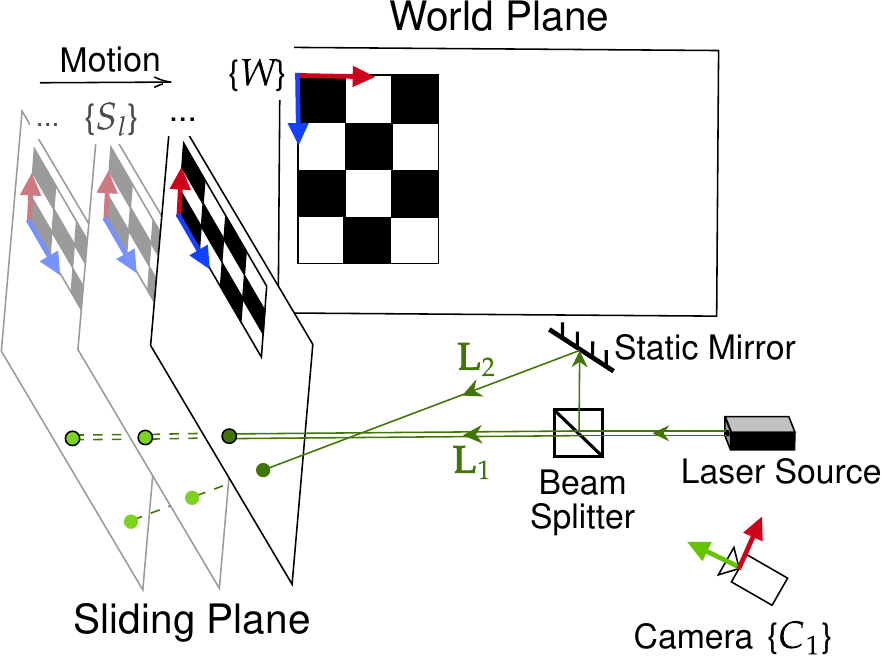}\label{fig:meas_incident_beam_setup}} \hspace*{.1in}
    \subfigure[]{\includegraphics[height=1.26in]{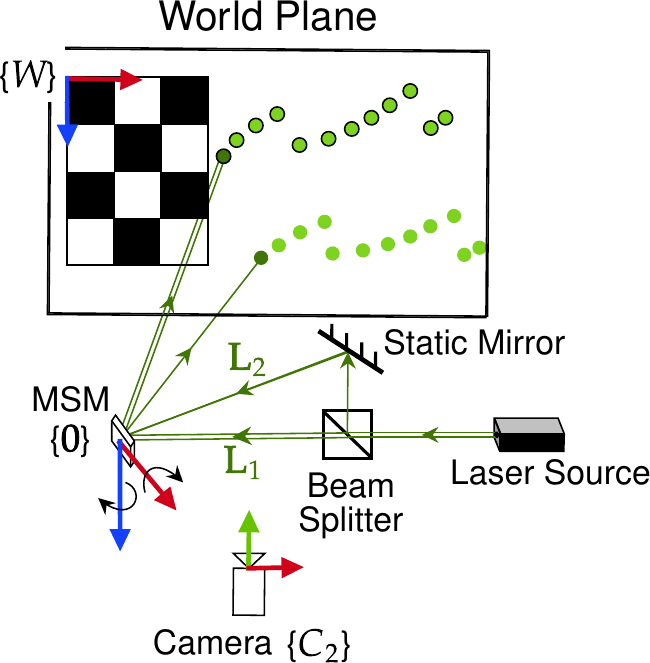}\label{fig:mirror_pose_est_setup}}
    \caption{(a) Schematics of the MSM (best viewed in color), components of the fast axis, slow axis and out-of-plane translation are colored in red, blue and green, respectively. (b) Incident beam estimation setup. (c) Mirror pose estimation setup.} 
\end{figure}

To address the challenge, we present a new calibration system and the corresponding algorithm design for the dynamic 3-DoF MSM system, which consists of a soft-hinged MSM with a triaxial Hall effect feedback sensor (Fig.~\ref{fig:mirror}) ~\cite{duan'19,duan'21}. Our contributions are threefold. First, we design a low-cost minimal 2-laser beam approach to reduce hardware cost (Figs.~\ref{fig:meas_incident_beam_setup}) and \ref{fig:mirror_pose_est_setup}). With the help of two planar calibration boards, the design can accurately estimate the 3-DoF MSM pose without using linear stages for precise optical alignments. Second, we derive calibration algorithms that build on a factor graph optimization framework that incorporates the reflection principle and conduct error analysis on the method. Third, we develop a self-synchronizing calibration scheme to establish the nonlinear mapping between Hall sensor readings and MSM poses. We have developed and implemented the entire system and algorithms. The results show that we can achieve an MSM poses estimation accuracy of $0.020^\circ$ with a standard deviation of $0.011^\circ$.

\section{Related Works} \label{sec:related}
Calibration is essential in the development and maintenance of a robotic system \cite{Roth'87,Verma'19}. Mechanism calibration and sensor calibration are two common types. The calibration of the robot mechanism focuses on estimating the kinematic or inertial parameters of the robot from actuator input and sensor measurements \cite{Mooring'91,Hollerbach'08}. Sensor calibration focuses on estimating the sensing model parameters from sensor measurements. Our MSM calibration is a combination of a mechanism calibration between the 3-DoF MSM poses and camera measurements and a sensor calibration between MSM poses and the triaxial Hall effect sensor measurements.

The mechanism calibration of MSM poses is related to manipulator calibration \cite{Nubiola'13} and hand-eye calibration as the mirror plane can be seen as an end-effector. While common practices of attaching markers to the end-effector for pose estimation are applicable for regular-sized static mirrors \cite{Chou'17, Lebraly'10}, they are not suitable for MSM due to its small size and dynamic scanning nature, attaching markers on the MSM will result in a change of scanning dynamics and deviated scanning poses. Similarly, the estimation methods that utilize real-virtual point constraints for regular-sized static mirrors proposed in \cite{Rodrigues'10,Ali'11,Yan'13,Takahashi'16} are impractical because the requirement of observing the points on objects and their virtual counterparts in mirror cannot be satisfied during fast MSM scanning. To measure MSM poses while accommodating the small size and dynamic scanning constraints, stroboscopic interferometer and position-sensitive detector (PSD) based methods have been investigated in the existing literature. The stroboscopic interferometer incorporates a periodically pulsed light source to illuminate the MSM at a specific scanning phase and estimates 3-DoF MSM poses from the interferometric images \cite{Rembe'01-0, Rembe'01-1}. The stroboscopic interferometer setup proposed by Rembe \etal has been shown to be capable of measuring dynamic MSM with up to $\mu m$ out-of-plane translation and $\pm 12^\circ$ rotations \cite{Rembe'01-1}. Although a stroboscopic interferometer provides superior measurement accuracy, its limited measurement range and costly complicated setup obstruct its applications. PSD-based methods estimate 2-DoF MSM rotations by tracking a reflection point of an incident laser beam on the PSD \cite{Schroedter'18, Xia'17}. Recent research focuses on improving the accuracy and range of measurements. In \cite{YOO'19}, Yoo \etal proposed a PSD-based MSM test bench with 0.026$^\circ$ accuracy in the 15$^\circ$ MSM scanning range. Baier \etal incorporated a PSD camera with a ray-trace shifting technique into their MSM test bench and achieved a measurement uncertainty of less than 1\% in the 47$^\circ$ MSM scanning range \cite{Baier'22}. These existing PSD based methods assume a precise alignment of the incident laser beam and the rotation center of the MSM due to their limitation in differentiating MSM translational motion with rotational motion, which impacts their accuracy when MSM out-of-plane translation is non-negligible or the incident beam fails to align with the mirror rotation center. Inspired by these existing works, our MSM mechanism calibration measures dynamic 3-DoF MSM poses by tracking the reflection of multiple incident laser beams generated by a strobe light with a camera.


Time offset estimation is required when the temporal misalignment in calibration measurements is not negligible, which is common when sensors have different clocks and sampling rates \cite{Enebuse'21}. In \cite{Kelly'14,Taylor'16}, the time offset is estimated by aligning the rotational changes measured by the sensors. Xia \etal show the independent estimations of time offset and the linear relationship between the motion of the MSM and the acoustic feedback in \cite{Xia'17}. A joint estimate of the time offset and other intrinsic and extrinsic parameters is preferred when the sensors do not have common measurements or follow a simple linear relationship \cite{Li'14,Rehder'16,Corte'19}. Building on existing methods, we propose an MSM calibration approach that jointly estimates time offset and model parameters to incorporate the nonlinear relationship between MSM motion and Hall effect sensor feedback. 

\section{Calibration System Design} \label{sec:Calib_system}
The MSM mechanism is reviewed before we elaborate the calibration procedure and the design of the rig.

\subsection{MSM Mechanism Review}

Fig.~\ref{fig:mirror} illustrates the mechanical structure of our 2-axis MSM that is detailed in our previous work~\cite{duan'19,duan'21}. Each mirror axis has a pair of soft hinges which form a gimbal structure to support the inner and the middle mirror frames. When currents flow through actuation coils of each axis, a magnetic force is generated and applied to the corresponding actuation magnets to rotate the mirror frame around the hinge pair. A sensing magnet is mounted on the back of the mirror plate. Therefore, the MSM motion changes the sensing magnet's magnetic field, which is perceived by a Hall effect sensor mounted on the fixed base plate.

The mirror scanning motion is actuated by applying sine wave-shaped alternating currents to the coils. For each scanning axis, the maximum scanning angle is achieved when the frequency of the input sine wave signal matches the resonance frequency of the MSM mechanism, which is the resonant scanning mode of the mirror. The mirror motion has 3 DoFs which include two rotational motions (one is fast and the other is slow) and out-of-plane translation because the soft hinges are made of polymeric materials. Before detailing the MSM calibration principle, we introduce common notations as follows.

\subsection{\textit{Nomenclature}}
All 3D coordinate systems or frames are right handed and Euclidean unless specified. $\mathbb{P}^2$ and $\mathbb{P}^3$ are 2D and 3D projective coordinate systems, respectively. $\mathbb{S}^2$ is the unit 2-sphere in the 3D Euclidean coordinate system, $\mathrm{T}_{\mathbf{v}}\mathbb{S}^2$ is the tangent space at the point $\mathbf{v}\in\mathbb{S}^2$. $[\cdot]_\times$ demotes skew-symmetric matrix.
\begin{itemize}
\item[$\{\mathbf{0}\}$] represents the MSM home frame, which is a fixed 3D system defined by the MSM home position. Its origin is at the MSM rotation center. Its Z-axis is parallel to the MSM normal vector. Its X-axis is parallel to the mirror fast axis.
\item[$\{W\}$] is a fixed 3D frame defined by a fixed world plane $\pi_\mathrm{W}$. Its origin is in the upper left corner of the checkerboard pattern in $\pi_\mathrm{W}$. Its Z-axis is perpendicular to $\pi_\mathrm{W}$ and points inward. Its X-axis is parallel to the horizontal direction of the checkerboard pattern.
\item[$\mathbf{\tilde{x}}$] is a point in the image, $\tilde{\mathbf{x}} \in \mathbb{R}^2$. Its homogeneous counterpart is $\mathbf{x} = [\mathbf{\tilde{x}}^\mathsf{T} \textrm{  }  1]^\mathsf{T} \in \mathbb{P}^2$.
\item[$\mathrm{\tilde{X}}$] is a point in the 3D Euclidean space, $\tilde{\mathrm{X}} \in \mathbb{R}^3$. Its homogeneous counterpart is $\mathrm{X} = [\mathrm{\tilde{X}}^\mathsf{T} \textrm{  }  1]^\mathsf{T} \in \mathbb{P}^3$.
\item[$\pi$] is a plane. $\pi = [\mathbf{n}^\mathsf{T} \textrm{  } d]^\mathsf{T}$, $\mathbf{n}\in\mathbb{S}^2$ is its unit length normal vector and $d$ is its distance to the origin. 
\item[$\mathbf{L}$] is a 3D line. $\mathbf{L} = [[\mathbf{v}]_\times \textrm{  }  \mathbf{m}]$, $\mathbf{v}\in\mathbb{S}^2$ is its unit length direction vector and $\mathbf{m}\in\mathrm{T}_{\mathbf{v}}\mathbb{S}^2$ is its moment vector at coordinate \plucker \cite{Mason'01,Taylor'94}. 
\item[$\mathrm{B}$] is a triaxial Hall effect sensor measurement vector of the magnetic field,  $\mathrm{B} = [b_x \textrm{  } b_y \textrm{  } b_z]^\mathsf{T}\in\mathbb{R}^3$.
\end{itemize}

 We use the left superscript to denote the coordinate system of an object; $^{W}\mathrm{\tilde{X}}$ is a point in the coordinate system $\{\textit{W}\}$. Variables without a specified coordinate are defaulted to $\{\textit{W}\}$. 

\subsection{Calibration Principle}\label{sc:cali_principle}
Recall that an MSM has a compact size and driving frequency-dependent working range; we cannot directly attach markers to it. Instead, we estimate MSM poses during resonant scanning by observing the reflected pulse laser dot positions on a world plane $\pi_\mathrm{W}$. Let us explain the working principle.

\begin{figure}[!htbp]
    \vspace*{-0.1 in}
    \centering
    \subfigure[]{\includegraphics[width=1.55in ]{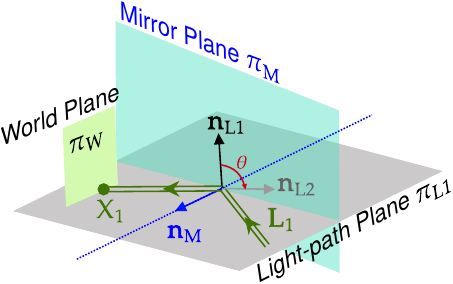}\label{fig:WorkingPrinciple-normal}} \hspace{.02in}
    \subfigure[]{\includegraphics[width=1.5in]{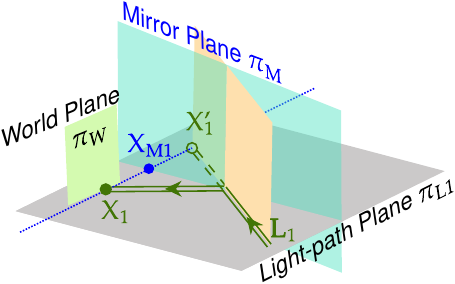}\label{fig:WorkingPrinciple-point}}
    \caption{Mirror pose estimation principle (best viewed in color): (a) Mirror normal $\mathbf{n}_\mathrm{M}$ is estimated from light-path plane normal vectors $\mathbf{n}_\mathrm{L1}$ and $\mathbf{n}_\mathrm{L2}$, $\theta$ is the spanning angle between them. (b) A point $\mathrm{X}_\mathrm{M1}$ on mirror plane is estimated from the real-virtual points $\mathrm{X}_{1}$ and $\mathrm{X}'_{1}$.} \label{fig:workingprinciple}
\end{figure}

Fig.~\ref{fig:WorkingPrinciple-normal} shows that the incident laser beam $\mathbf{L}_1$ and its reflected laser dot $\mathrm{X}_{1}$ on $\pi_\mathrm{W}$
define a light-path plane $\pi_\mathrm{L1}$, which is perpendicular to the mirror plane $\pi_\mathrm{M}$. Therefore, its normal vector $\mathbf{n}_\mathrm{L1}$ must also be perpendicular to the mirror normal $\mathbf{n}_\mathrm{M}$, this forms a single DoF constraint. By including another incident laser beam, we can obtain the new normal vector $\mathbf{n}_\mathrm{L2}$ of its light-path plane in a similar way. We maintain $\mathbf{n}_\mathrm{L2} \notparallel \mathbf{n}_\mathrm{L1}$ when choosing the second incident laser beam. Therefore, the normal vector of the two DoF mirror plane $\mathbf{n}_\mathrm{M}$ can be derived as follows,
$\mathbf{n}_\mathrm{M} = \mathbf{n}_\mathrm{L1} \times \mathbf{n}_\mathrm{L2},$ where `$\times$' means cross product. The spanning angle $\theta$ between $\mathbf{n}_\mathrm{L1}$ and $\mathbf{n}_\mathrm{L2}$ is a control variable, and we will discuss its effect on the uncertainty of the estimation in Sec. \ref{Mirror_Est_Result}.

Once the 2-DoF mirror plane normal is determined, the last DoF of the MSM pose can be determined by identifying any point on the mirror plane. Here, we identify the middle point $\mathrm{X}_\mathrm{M1}$ between the reflected laser dot $\mathrm{X}_{1}$ and its reflection point $\mathrm{X}'_{1}$. We know the line $\overline{\mathrm{X}_{1}\mathrm{X}'_{1}} \parallel \mathbf{n}_\mathrm{M}$. Therefore, the line $\overline{\mathrm{X}_{1}\mathrm{X}'_{1}}$ is uniquely defined because we know $\mathrm{X}_{1}$ and $\mathbf{n}_\mathrm{M}$. As shown in Fig.~\ref{fig:WorkingPrinciple-point}, the intersection of the line $\overline{\mathrm{X}_{1}\mathrm{X}'_{1}}$ and the incident beam line $\mathbf{L}_1$ is $\mathrm{X}'_{1}$. With $\mathbf{n}_\mathrm{M}$ and $\mathrm{X}_\mathrm{M1}=\frac{1}{2}(\mathrm{X}_{1}+\mathrm{X}'_{1})$, the mirror plane $\pi_\mathrm{M}$ is uniquely determined. In summary, with two incident laser beams $\mathbf{L}_1 \notparallel \mathbf{L}_2$ and their corresponding observation points $\mathrm{X}_{1}$ and $\mathrm{X}_{2}$, the mirror plane $\pi_\mathrm{M}$ is uniquely defined. 

\subsection{Calibration Rig Design and Procedure}
To obtain two pairs of non-parallel laser beams and their reflected laser dots, we employ a beam splitter to generate two laser beams from a pulse laser source and a camera to observe the reflected laser dots positions. This leads to a two-step process described by Figs.~\ref{fig:meas_incident_beam_setup} and \ref{fig:mirror_pose_est_setup}.

The first step is to obtain the 3D line parameters of the incident beams. Fig.~\ref{fig:meas_incident_beam_setup} shows the setup where a fixed camera observes the sliding plane $\pi_\mathrm{S}$ and the fixed world plane $\pi_\mathrm{W}$.  The camera is placed with a good view of the sliding planes.
The corresponding camera coordinate system is defined as $C_1$. The MSM is not mounted in this step to allow the two incident beams to project points directly onto $\pi_\mathrm{S}$. We track their projected laser dots on a sliding plane, since laser beams are not directly visible in the camera image.
When we move $\pi_\mathrm{S}$ closer to the laser source, the positions of the laser points on $\pi_\mathrm{S}$ change with motion. The sliding plane coordinate system $\{\textit{S}_l\}$ with its $l$-th pose is defined with respect to its checkerboard, similar to how $\{\textit{W}\}$ is defined. To reconstruct the incident beams, from image $I_{l}$ we extract the laser points $\mathbf{x}_{i,l}$ of the $i$-th incident beam and the checkerboard corner points $\mathbf{x}_{s,l}$ and $\mathbf{x}_{w,l}$, where $s$ and $w$ are index variables for the $s$-th and the $w$-th corner points on the sliding plane and world plane, respectively.

At the end of the step, before the movable part of the MSM (i.e. the top frame of the mirror in Fig.~\ref{fig:mirror}) is assembled, we also collect background magnetic field measurements $\mathrm{B}_{b}$ which include periodic background noises generated by actuator coils. 
We use function generators to drive the actuation coils with the sine wave signals that excite resonance mirror scanning, and record background measurements $\mathrm{B}_{b}$ from the Hall effect sensor. We will show how to use $\mathrm{B}_{b}$ to cancel background noise later in the paper. After this step, the MSM is fully assembled to measure the actual magnetic field $\mathrm{B}_{a}$ during mirror scanning.

Fig.~\ref{fig:mirror_pose_est_setup} shows the second step in estimating the mirror pose, where the camera aims at $\pi_\mathrm{W}$ and the MSM is mounted to reflect incident beams to project points onto $\pi_\mathrm{W}$. The camera pose is adjusted to have a good view of $\pi_\mathrm{W}$ with its camera coordinate system defined as $C_2$. Note that incident beams maintain the same configuration as in the last step. 

The synchronized pulse laser and the mirror scanning signals create a pair of dotted scanning patterns from the two incident beams. The $k$-th image $I_{k}$ captures the checkerboard corner points $\mathbf{x}_{w,k}$ on $\pi_\mathrm{W}$ and the reflected laser points $\mathbf{x}_{i,j}$ of the incident beam $i$-th triggered at time $t_j$. 

In image processing, we apply color thresholding to extract laser dots from images. For each laser dot, the mean position of extracted pixels is used to represent its 2D position in image. By the central limit theorem, the $i$-th laser dot position $\mathbf{x}_i$ follows a Gaussian distribution $\mathcal{N}(0,\Sigma_{\mathbf{x}i})$, where $\Sigma_{\mathbf{x}i} = \frac{\Sigma_\mathrm{Pi}}{N}$ is the covariance matrix of $\mathbf{x}_i$, $\Sigma_\mathrm{Pi}$ is the covariance matrix of the 2D pixel positions, and $N$ is the number of extracted pixels. 

\subsection{Signal Synchronization and Sparse Signal Triggering}

\begin{figure}[!htbp]
    \centering
    \subfigure[]{\includegraphics[height=1.1in]{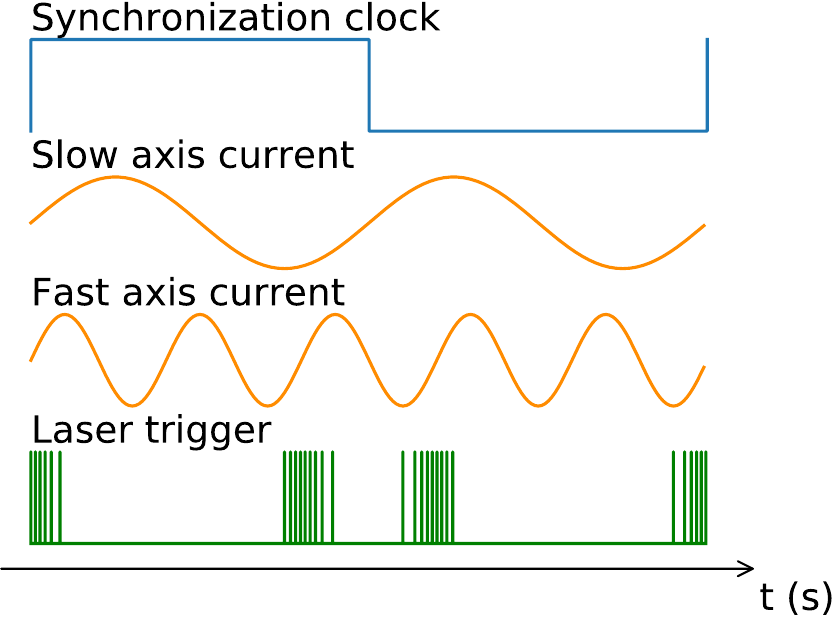}\label{fig:4signals}}
    \begin{minipage}{1.5in}
    \vspace*{-.8in}
    \centering
    \subfigure[]{\includegraphics[width=1.2in]{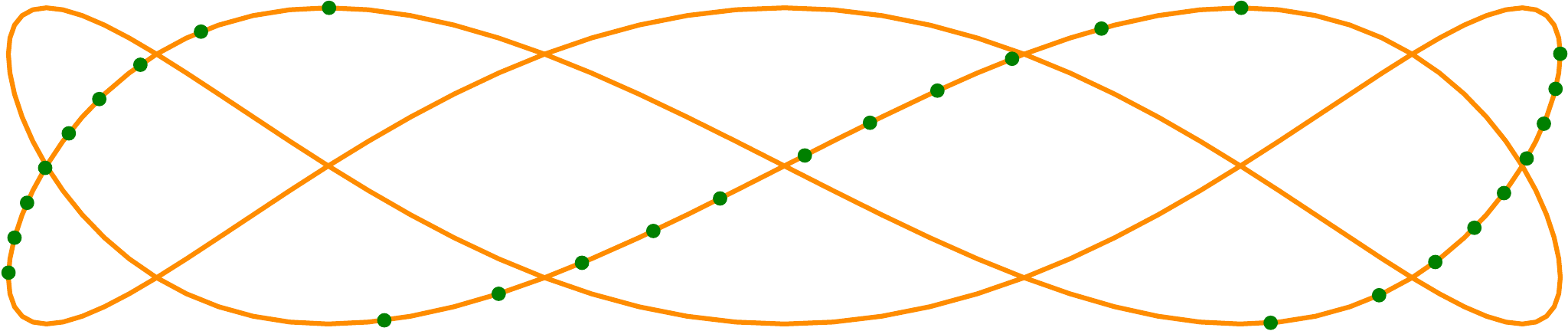}\label{fig:laser_pattern_ideal}} \\
    \subfigure[]{\includegraphics[width=1.2in]{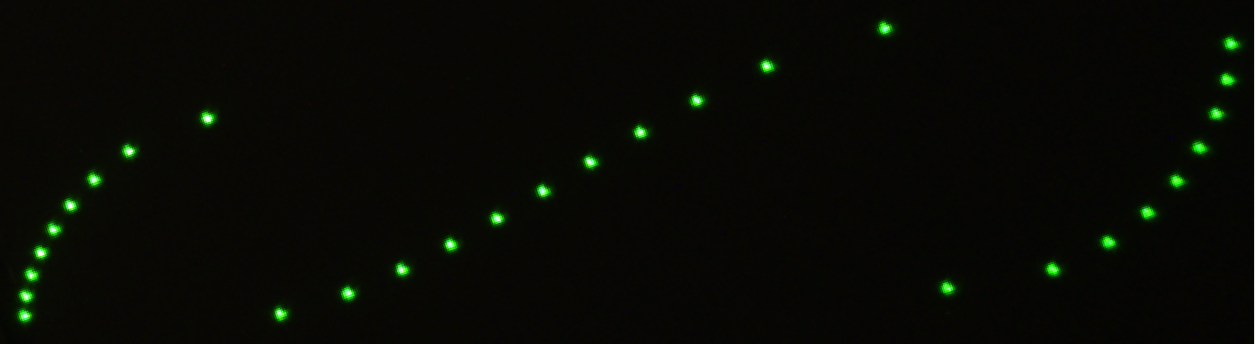}\label{fig:laser_pattern_real}} 
    \end{minipage}
    \caption{(a) One cycle of the 4 signals generated. (b) Expected scanning pattern where the orange line is the ideal laser dot trajectory and green dots are the locations illuminated by the laser pulses. (c) The observed dot pattern in the image.} \label{fig:electronic-components-signals}
\end{figure}

To capture dynamic mirror motion and reduce motion blur caused by mirror scanning, we use a pulsed laser with a 15 ns pulse width as our strobe light source, which also frees the camera from triggering or synchronizing.
To establish the correspondence between laser dot positions, mirror driving signals, and Hall sensor readings, we use a function generator (FG) to provide 4 synchronized signals (Fig.~\ref{fig:4signals}) that include a pulse signal to trigger the laser source, two sine wave driving signals to activate coils in the MSM, and a clock signal to align with Hall effect sensor interrupt signals generated by the microcontroller unit (MCU).

Because cluttered laser dots in an image may lead to incorrect dot center estimation, we generate the laser trigger signals according to the mirror motion, which makes the laser dots sparsely spaced. Since the rotation angle is nearly linear to the driving current of the mirror, we can match the trigger signal with the driving sine waves to ensure the sparsity of the laser dots. Fig.~\ref{fig:electronic-components-signals} illustrates an example where the two sine-wave signals drive the corresponding mirror axis. To avoid cluttering, the laser pulse is triggered when both sine waves have a positive gradient and their vertical distance in signal space (Fig.~\ref{fig:laser_pattern_ideal}) is constant. 

\section{Problem Formulation} \label{sec:Problem Formulation}

We have the following assumptions:
\begin{itemize}
\item[a.1] The camera is pre-calibrated which means known intrinsic parameters with lens distortion removed.
\item[a.2] The MSM scanning pattern is repeatable given the same input current sequence.
\end{itemize}

Mirror calibration is a two-step process. The first step is a mirror pose estimation problem.
\begin{Def}[Mirror Pose Estimation]\label{def:mirror_pose}
Given the observation points of the two incident laser beams $\mathbf{x}_{i,l}$ and their reflected laser points $\mathbf{x}_{i,j}$ in their respective image coordinates and checkerboard points $\mathbf{x}_{s,l}$, $\mathbf{x}_{w,l}$ and $\mathbf{x}_{w,k}$ in the image, estimate the mirror planes ${}^\mathbf{0}\tilde{\pi}_{\mathrm{Mj}}$. 
\end{Def}
\begin{figure}[!ht]
    \centering
    \includegraphics[width=3.2 in]{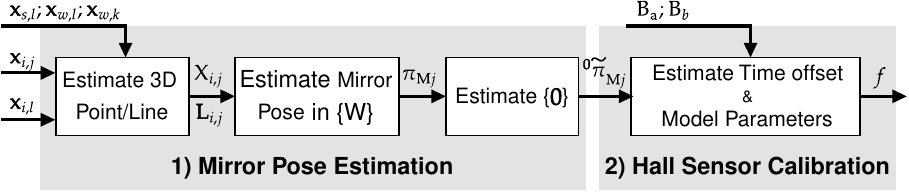}
    \caption{Calibration block diagram.} \label{fig:procedure}
\end{figure}
The second step is to model and calibrate the mapping between Hall sensor readings and mirror poses.
\begin{Def}[Hall Sensor Calibration]
Given a sequence of mirror planes ${}^\mathbf{0}\tilde{\pi}_{Mj}$, sequence of background magnetic field measurements $\mathrm{B}_{b}$ and actual measurements $\mathrm{B}_{a}$ from the Hall sensor, estimate the time offset $\delta t$ and the parameters of the model $f:\mathbb{R}^3\rightarrow\mathbb{R}^3$ that maps Hall sensor readings to mirror poses.
\end{Def}

\section{Calibration Algorithm} \label{sec:EstMethod}

The calibration pipeline is shown in \autoref{fig:procedure}. We start with mirror pose estimation.

\subsection{Mirror Pose Estimation}\label{ssc:MirrorPoseEst}
For simplicity, we omit the index subscript $j$ for the variables associated with time $t_j$ before Sec. \ref{sssc::estimte_0}. In other words, the points $\mathbf{x}_{i,j}$ and $\mathrm{X}_{i,j}$, the lines $\mathbf{L}_{i,j}$ and the planes $\pi_{\mathrm{Mj}}$ will be noted as $\mathbf{x}_{i}$, $\mathrm{X}_{i}$, $\mathbf{L}_{i}$, and $\pi_\mathrm{M}$, respectively, in Secs.~\ref{subsec:3D-Point-Line} and \ref{subsec:MirrorPose}.
\subsubsection{Estimate 3D Point/Line}\label{subsec:3D-Point-Line}
Because the mirror pose is estimated from points in camera image, let us first introduce the camera projection model and then explain how to obtain the transformation between the camera coordinate systems and $\{W\}$ and $\{S\}$ defined by the planes $\pi_\mathrm{W}$ and $\pi_\mathrm{S}$.

According to \cite{hartley'04}, a 3D point $\mathrm{X}$ in world coordinate $\{W\}$ and its counterpart $\mathbf{x}$ in camera image satisfies
\begin{equation} \label{eq:PnPConstrain}
\mathbf{x} = \lambda\mathbf{K}[\mathbf{R} \textrm{  } \mathbf{t}]\mathrm{X}.
\end{equation}
Here $\lambda$ is a scaling factor, $\mathbf{K}$ is the intrinsic matrix of the camera, $\mathbf{R}$ and $\mathbf{t}$ are rotation and translation components of the transformation matrix ${}_{W}^{C}\mathbf{T} = \begin{bmatrix}\mathbf{R} & \mathbf{t} \\ \mathbf{0} & 1\end{bmatrix}$, which transform from coordinate of the world plane $\{W\}$ to coordinate of the camera $\{C\}$. ${}_{W}^{C}\mathbf{T}$ is estimated by solving the perspective n-point problem (PnP) \cite{Li'12}
with $\mathbf{K}$  from camera calibration and checkerboard points.

In the setup shown in Fig.~\ref{fig:meas_incident_beam_setup}, the transformations ${}_{W}^{C_1}\mathbf{T}$ and ${}_{S_l}^{C_1}\mathbf{T}$ between the camera coordinate system $\{C_1\}$, $\{W\}$ and $\{S_l\}$ are obtained from the checkerboard corner points $\mathbf{x}_{s,l}$, $\mathbf{x}_{w,l}$ and their corresponding 3D planar checkerboard corner points by solving the PnP problem. Similarly, for the camera coordinate system $\{C_2\}$ in the second step (Fig.~\ref{fig:mirror_pose_est_setup}), ${}_{W}^{C_2}\mathbf{T}$ is solved with PnP using checkerboard corner points $\mathbf{x}_{w,k}$ and their 3D counterparts. Because the grid size of the checkerboard pattern is known, the true scale is recovered in the process.

3D point $\mathrm{X}_{i}$ on the world plane can be derived from its image counterpart $\mathbf{x}_{i}$ with the transformation matrix ${}_{\textit{W}}^{\textit{C}_2}\mathbf{T}$ estimated from PnP, as $\mathrm{\tilde{X}}_{i} = \mathbf{R}^\mathsf{T}(\frac{1}{\lambda}\mathbf{K}^{-1}\mathbf{x}_{i}- \mathbf{t})$, here $\mathbf{R}$ and $\mathbf{t}$ are components from ${}_{\textit{W}}^{\textit{C}_2}\mathbf{T}$ and $\lambda = \frac{[\mathbf{R}^\mathsf{T}\mathbf{K}^{-1}]_3}{[\mathbf{R}^\mathsf{T}\mathbf{t}]_3} \mathbf{x}_{i}$, where $[\cdot]_3$ denotes the third row of a vector or matrix. 

The points in the sliding plane $\pi_\mathrm{S}$ share the same derivation as the points in $\pi_\mathrm{W}$. Therefore, we can obtain the observation points of all incident laser beams ${}^{S_{l}}\mathrm{X}_{i,l}$ from $\mathbf{x}_{i,l}$. We transform the points in $\{\textit{S}_l\}$ to $\{\textit{W}\}$ as $\mathrm{X}_{i,l} = {}_{C_1}^{W}\mathbf{T}{}_{S_{l}}^{C_1}\mathbf{T}{}^{S_{l}}\mathrm{X}_{i,l}$, where the transformation matrix ${}_{C_1}^{W}\mathbf{T}$ and ${}_{S_{l}}^{C_1}\mathbf{T}$ is obtained by solving the PnP problem. 

We represent a 3D line with \plucker coordinates. The incident laser beam $\mathbf{L}_{i} = [[\mathbf{v}_i]_\times \textrm{  }  \mathbf{m}_{i}]$ is formed by the direction vector $\mathbf{v}_i$ and the moment vector $\mathbf{m}_i$.
The direction vector $\mathbf{v}_i$ can be estimated from the points in the incident laser beam by the principal component analysis (PCA) as $(\mathbf{\tilde{X}}_\mathrm{Li}-\mathrm{\bar{X}}_\mathrm{Li})^\mathsf{T} = \mathbf{U}\mathbf{S}\mathbf{V}^\mathsf{T}$, where $\mathbf{\tilde{X}}_\mathrm{Li} = \begin{bmatrix} \ldots & \mathrm{\tilde{X}}_{i,l} & \ldots \end{bmatrix}$ are the laser dots observed in the sliding plane as shown in Fig.~\ref{fig:meas_incident_beam_setup}, $\mathrm{\bar{X}}_\mathrm{Li}$ is the mean of the row of $\mathbf{\tilde{X}}_\mathrm{Li}$. The first principal component of $\mathbf{V}$ is the direction vector $\mathbf{v}_i$ of $\mathbf{L}_{i}$. 
The moment vector $\mathbf{m}_i$ is given by $\mathbf{m}_{i} = \mathbf{v}_i \times \mathrm{\bar{X}}_\mathrm{Li}$ following the conventions of \cite{Mason'01}.

Any point $\mathrm{X}_{i,l}$ on $\mathbf{L}_{i}$ satisfies $\mathbf{v}_i \times \mathrm{\tilde{X}}_{i,l}+\mathbf{m}_i = \mathbf{0}$ and  $\mathbf{X}_\mathrm{Li} = [\mathbf{\tilde{X}}_\mathrm{Li}^{\mathsf{T}} ~~ \mathbf{1}]^\mathsf{T}$, a line constrain can be formulated as
\begin{equation} \label{eq:LineConstrain}
\mathbf{L}_{i} \mathbf{X}_\mathrm{Li} = \mathbf{0}.
\end{equation}

For any point $\mathrm{X}_{i}$ not on $\mathbf{L}_{i}$, the normal vector of the light path plane they form is given by $\mathbf{n}_\mathrm{Li} = \mathbf{v}_i \times (\mathrm{\tilde{X}}_{i}-\mathbf{\bar{X}}_{Li}) = \mathbf{L}_{i} \mathrm{X}_{i}$.

\subsubsection{Estimate Mirror Pose in $\{\textit{W}\}$} \label{subsec:MirrorPose}
As discussed in Sec.~\ref{sc:cali_principle}, mirror plane $\pi_\mathrm{M}$ is calculated from incident beams $\mathbf{L}_i$ and their reflected laser dots $\mathrm{X}_i$.

The mirror plane $\pi_\mathrm{M}$ is perpendicular to the light path planes $\pi_\mathrm{Li}$ means that the mirror normal $\mathbf{n}_\mathrm{M}$ and the normal light path planes $\mathbf{n}_\mathrm{Li}$ are also perpendicular. Therefore, $\mathbf{n}_\mathrm{M}$ can be solved from $\mathbf{N}_\mathrm{L}^{\mathsf{T}}\mathbf{n}_\mathrm{M} = \mathbf{0}$, where $\mathbf{N}_\mathrm{L} = \begin{bmatrix} \ldots & \mathbf{n}_\mathrm{Li} & \ldots \end{bmatrix}$ contains all the normals in the light-path plane.

As shown in Fig.~\ref{fig:WorkingPrinciple-point}, for a reflected laser dot $\mathrm{X}_i$, its virtual counterpart $\mathrm{X}_i'$ lies on the extension of the incident beam $\mathbf{L}_i$. $\mathrm{X}_i'$ can be derived from the reflection transformation as $\mathrm{\tilde{X}}_i' = \mathrm{\tilde{X}}_i-2(\mathbf{n}_\mathrm{M}^\mathsf{T}\mathrm{\tilde{X}}_i-d_\mathrm{M})\mathbf{n}_\mathrm{M}$ \cite{Rodrigues'10}. Therefore a reflection constrain can be formulated as
\begin{equation} \label{eq:ReflectionConstrain}
    \mathbf{L}_i\mathbf{H}\mathrm{X}_i = \mathbf{0}
\end{equation}
where $\mathbf{H} = \begin{bmatrix} \mathbf{I}-2\mathbf{n}_\mathrm{M}\mathbf{n}_\mathrm{M}^\mathsf{T} & -2d_\mathrm{M}\mathbf{n}_\mathrm{M} \\ \mathbf{0} & 1\end{bmatrix}$ is the reflection transformation matrix. 

To obtain the optimal mirror estimation results from the initial solutions solved using (\ref{eq:PnPConstrain}), (\ref{eq:LineConstrain}) and (\ref{eq:ReflectionConstrain}), we formulate a Maximum Likelihood Estimation (MLE) problem that jointly refines the parameters from measurements in camera images.


During optimization, the variables are represented in their minimum parameterization to improve computation efficiency. With the logarithmic maps $\Log_\mathbf{R}:SO(3)\rightarrow\mathbb{R}^3$ and $\Log_\mathbf{q}:\mathbb{S}^3\rightarrow\mathbb{R}^3$ defined in \cite{Joan'18}, the transformation matrix $\mathbf{T}$ is represented as $\mathrm{\tilde{T}} = [\Log_\mathbf{R}(\mathbf{R})^\mathsf{T} \ \mathbf{t}^\mathsf{T}]^\mathsf{T}\in\mathbb{R}^6$, the plane $\pi$ is represented as to $\tilde{\pi} = \Log_\mathbf{q}(\frac{\pi}{\|\pi\|})\in\mathbb{R}^3$ \cite{Kaess'15},
the 3D line $\mathbf{L}$ is mapped to $\mathrm{\tilde{L}} = [\Log_\mathbf{R}(\mathbf{R}_L)^\mathsf{T} \ m]^\mathsf{T}\in\mathbb{R}^4$, where $m = \|\mathbf{m}\|$ and $\mathbf{R}_L = \begin{bmatrix}\mathbf{v} & \frac{\mathbf{m}}{m} & \mathbf{v} \times \frac{\mathbf{m}}{m} \end{bmatrix}$ \cite{Bartoli'05}.
The minimum parameterized variables are aggregated as $\mathcal{X}=[\mathcal{P}^\mathsf{T} \ \mathcal{L}^\mathsf{T} \ \mathcal{T}^\mathsf{T}]^\mathsf{T}$ to be optimized in MLE, where $\mathcal{P}=\begin{bmatrix} \ldots & \tilde{\pi}_{\mathrm{Mj}}^\mathsf{T} & \ldots \end{bmatrix}^\mathsf{T}$ are all the mirror planes, $\mathcal{L}=\begin{bmatrix} \ldots & \mathrm{\tilde{L}}_{i}^\mathsf{T} & \ldots \end{bmatrix}^\mathsf{T}$ are all the laser beams, and $\mathcal{T}=\begin{bmatrix} {}_{\textit{W}}^{\textit{C}_1}\mathrm{\tilde{T}}^\mathsf{T} & {}_{\textit{W}}^{\textit{C}_2}\mathrm{\tilde{T}}^\mathsf{T} & \ldots & {}_{\textit{S}_l}^{\textit{C}_1}\mathrm{\tilde{T}}^\mathsf{T} & \ldots \end{bmatrix}^\mathsf{T}$ are all the transformations between camera, the world plane and the sliding plane.

\begin{figure}[ht]
    \centering
    \includegraphics[width=3 in]{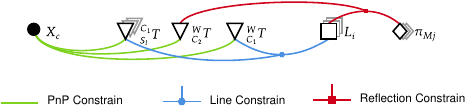}
    \caption{Factor graph illustration of the MLE problem.} \label{fig:FactorGraph}
\end{figure}

The cost function of the MLE problem is formulated as the reprojection errors of images points, and it has three components, their detailed derivations are included in the appendix. The first component $C_P$ (green edges in Fig.~\ref{fig:FactorGraph}) is from the checkerboard corner points observed in the calibration process (Fig.~\ref{fig:meas_incident_beam_setup} and Fig.~\ref{fig:mirror_pose_est_setup}). $C_P(\mathcal{X})$ is defined as
\begin{equation}\label{eq:Proj-Cost}
C_P(\mathcal{X}) =
\sum\|\mathbf{x}-f_{P}(\mathrm{\tilde{T}},\mathrm{X})\|_\Sigma^2
\end{equation}
where $f_{P}$ is the PnP constraint derived from (\ref{eq:PnPConstrain}). Here $(\mathbf{x},\mathrm{\tilde{T}},\mathrm{X}) \in \{(\mathbf{x}_{w,l},{}_{\textit{W}}^{\textit{C}_1}\mathrm{\tilde{T}},\mathrm{X}_c), (\mathbf{x}_{w,k},{}_{\textit{W}}^{\textit{C}_2}\mathrm{\tilde{T}},\mathrm{X}_c), (\mathbf{x}_{w,l},{}_{\textit{S}_l}^{\textit{C}_1}\mathrm{\tilde{T}},\mathrm{X}_c)\}$. $\mathrm{X}_c$ is the known 3D checkerboard points on the sliding plane and world plane predefined by the checkerboard pattern. $\|\cdot\|_\Sigma$ denotes the Mahalanobis distance.
The second cost function component $C_L$ (blue edges in Fig.~\ref{fig:FactorGraph}) is from the laser beam observation points shown in Fig.~\ref{fig:meas_incident_beam_setup}, $C_L(\mathcal{X})$ is defined as
\begin{equation}\label{eq:Line-Cost}
C_L(\mathcal{X}) = \sum_{i,l}\|\mathbf{x}_{i,l}-f_{L}({}_{\textit{S}_l}^{\textit{C}_1}\mathrm{\tilde{T}},\mathrm{\tilde{L}}_i,{}_{\textit{W}}^{\textit{S}_l}\mathrm{\tilde{T}})\|_\Sigma^2
\end{equation}
where $f_{L}$ is the line constraint derived from (\ref{eq:PnPConstrain}) and (\ref{eq:LineConstrain}) as the camera image projection of the intersecting point between laser beam $\mathrm{L}$ and sliding plane $\pi_{Sl}$.
The third component of the cost function $C_R$ (red edges in Fig.~\ref{fig:FactorGraph}) is from the reflected laser dots shown in Fig.~\ref{fig:mirror_pose_est_setup}. $C_R(\mathcal{X})$ is defined as
\begin{equation}\label{eq:Reflection-Cost}
C_R(\mathcal{X}) = \sum_{i,j}\|\mathbf{x}_{i,j}-f_{R}({}_{\textit{W}}^{\textit{C}_2}\mathrm{\tilde{T}},\mathrm{\tilde{L}}_i,\tilde{\pi}_\mathrm{Mj})\|_\Sigma^2
\end{equation}
where $f_{R}$ is the reflection constraint derived from (\ref{eq:PnPConstrain}) and (\ref{eq:ReflectionConstrain}) as the camera image projection of the intersecting point between laser beam $\mathrm{L}$ and reflected world plane $\pi_{W}$ with the reflection relationship defined by mirror plane $\pi_\mathrm{Mj}$.

The MLE of $\mathcal{X}$ is solved by minimizing
\begin{equation} \label{eq:MLE}
   \min_{\mathcal{X}^*} \ C_P(\mathcal{X}^*) + C_L(\mathcal{X}^*) + C_R(\mathcal{X}^*)
\end{equation}
using the Levenberg–Marquardt (LM) algorithm. And the uncertainty of $\mathcal{X}^*$ is given by
\begin{equation} \label{eq:MLE-Uncertainty}\resizebox{.91\hsize}{!}{$
\Sigma_{\mathcal{X}} = \left(
\sum\limits_{w,l}J_{w,l}^{\mathsf{T}}\Sigma^{-1}J_{w,l}+
\sum\limits_{w,k}J_{w,k}^{\mathsf{T}}\Sigma^{-1}J_{w,k}+
\sum\limits_{s,l}J_{s,l}^{\mathsf{T}}\Sigma^{-1}J_{s,l}+
\sum\limits_{i,l}J_{i,l}^{\mathsf{T}}\Sigma^{-1}J_{i,l}+
\sum\limits_{i,j}J_{i,j}^{\mathsf{T}}\Sigma^{-1}J_{i,j}
\right)^{-1}
$}\end{equation}
where 
$\begin{bmatrix} J_{w,l} \\ J_{w,k} \\ J_{s,l} \\ J_{i,l} \\ J_{i,j} \end{bmatrix} = 
\begin{bmatrix}
\frac{\partial f_{P}({}_{\textit{W}}^{\textit{C}_1}\mathrm{\tilde{T}}^*,\mathrm{X}_c)}{\partial\mathcal{X}} \vspace{.1em}\\ 
\frac{\partial f_{P}({}_{\textit{W}}^{\textit{C}_2}\mathrm{\tilde{T}}^*,\mathrm{X}_c)}{\partial\mathcal{X}} \vspace{.1em}\\ 
\frac{\partial f_{P}({}_{\textit{S}_l}^{\textit{C}_1}\mathrm{\tilde{T}^*},\mathrm{X}_c)}{\partial\mathcal{X}} \vspace{.1em}\\ 
\frac{\partial f_{L}({}_{\textit{S}_l}^{\textit{C}_1}\mathrm{\tilde{T}}^*,\mathrm{\tilde{L}}_i^*,{}_{\textit{W}}^{\textit{S}_l}\mathrm{\tilde{T}}^*)}{\partial\mathcal{X}} \vspace{.1em}\\ 
\frac{\partial f_{R}({}_{\textit{W}}^{\textit{C}_2}\mathrm{\tilde{T}}^*,\mathrm{\tilde{L}}_i^*,\tilde{\pi}_\mathrm{Mj}^*)}{\partial\mathcal{X}}
\end{bmatrix}$
are the Jacobians.  

To validate the parameters estimated from MLE, we use the Euclidean distance between the reflected laser dot observations $\mathbf{x}_{m,j}$ in testing set and the predicted projection of the reflected laser dot as our evaluation metric. Because the data used for parameter estimation are not overlapped with the testing data, we note the index variables $m\neq i \ \forall m,i$. Let $\mathbf{p}_{m,j} = [ (\mathcal{T}^*)^\mathsf{T},   (\tilde{\pi}_\mathrm{Mj}^*)^\mathsf{T},   \mathbf{x}_{m}^\mathsf{T}]^{\mathsf{T}}$ be the parameters we use for the prediction, where $\mathbf{x}_{m} = [\ldots \ \mathbf{x}_{m,l}^\mathsf{T} \ \ldots]^\mathsf{T}$. The prediction error is
\begin{equation} \label{eq:MLE-PredictionError}
\delta_{m,j} = \| \mathbf{x}_{\mathrm{m,j}}-f_{pred}(\mathbf{p}_{\mathrm{m,j}})\|_2
\end{equation}
where $\|\cdot\|_2$ is the L2 norm, $f_{pred}$ is the projection prediction function derived from (\ref{eq:PnPConstrain}), (\ref{eq:LineConstrain}) and (\ref{eq:ReflectionConstrain}). 
Under the Gaussian noise assumption, the variance of the prediction error is 
\begin{equation} \label{eq:MLE-PredictionError-Variance}
\sigma_{m,j}^2 = J_{\mathbf{x}}\Sigma_\mathbf{x}J_\mathbf{x}^{\mathsf{T}} + J_\mathbf{p}\Sigma_\mathbf{p}J_\mathbf{p}^{\mathsf{T}}
\end{equation}
where $J_\mathbf{x} = \frac{\partial \delta_{m,j}}{\partial \mathbf{x}_{m,j}}$ and $J_\mathbf{p} = \frac{\partial \delta_{m,j}}{\partial \mathbf{p}_{m,j}}$. $\Sigma_\mathbf{p} = \begin{bmatrix}\Sigma_{\mathcal{T}\pi} & \\ & \Sigma_{\mathbf{x}m}\end{bmatrix}$, where $\Sigma_{\mathcal{T}\pi}$ is a submatrix of $\Sigma_{\mathcal{X}}$ derived in \ref{eq:MLE-Uncertainty} and $\Sigma_{\mathbf{x}m} = \textrm{diag}(\ldots,\Sigma_{\mathbf{x}},\ldots)$ is a block diagonal matrix with all the covariance matrix of $\mathbf{x}_{m,l}$.

\subsubsection{Estimate $\{\mathbf{0}\}$}\label{sssc::estimte_0}
Frame transformation from $\{\textit{W}\}$ and $\{\mathbf{0}\}$ is to be estimated from the transient mirror poses $\pi_\mathrm{Mj}$. This is a necessary step because we need to extract the mirror poses in $\{\mathbf{0}\}$ before we can map them to the Hall sensor readings.

By definition, the X-axis of the mirror coordinate system is parallel to the mirror fast axis, which is perpendicular to all mirror normals during 1D fast axis scanning, this means that we can estimate its directional vector $\mathbf{e}_\mathrm{F}$ from
$\mathbf{N}_\mathrm{F}^\mathsf{T}\mathbf{e}_\mathrm{F} = \mathbf{0}$ 
where $\mathbf{N}_\mathrm{F} = \begin{bmatrix} \ldots & \mathbf{n}_{\mathrm{Mf}} & \ldots \end{bmatrix}$ contains all the mirror normals during 1D fast axis scanning. The Z axis of the mirror coordinate system is parallel to the normal vector of the neutral mirror position $\mathbf{n}_\mathrm{M0}$.
Therefore, the rotation matrix from $\{\mathbf{0}\}$ to $\{W\}$ is ${}_{\mathbf{0}}^{W}\mathbf{R} = \begin{bmatrix}\mathbf{e}_\mathrm{F} & \mathbf{n}_\mathrm{M0}\times\mathbf{e}_\mathrm{F} & \mathbf{n}_\mathrm{M0}\end{bmatrix}$. 

The frame $\{\mathbf{0}\}$'s origin $\mathrm{X}_\mathrm{O}$ is defined as the center of mirror rotation; in other words, it is the point that shares all mirror planes. Therefore, $\mathrm{X}_\mathrm{O}$ satisfies
\begin{equation} \label{eq:MirrorTranslation}
\Pi_\mathrm{M}^\mathsf{T}\mathrm{X}_\mathrm{O} = \mathbf{0}
\end{equation}
where $\Pi_\mathrm{M} = \begin{bmatrix} \ldots & \pi_\mathrm{Mj} & \ldots \end{bmatrix}$ contains all the mirror poses.

We can now derive the transformation matrix from $\{\textit{W}\}$ to $\{\mathbf{0}\}$ as ${}_{W}^{\mathbf{0}}\mathbf{T} =\begin{bmatrix}{}_{\mathbf{0}}^{W}\mathbf{R}^\mathsf{T} & -_{\mathbf{0}}^{W}\mathbf{R}^\mathsf{T}\mathrm{X}_\mathrm{O} \\ \mathbf{0} & 1\end{bmatrix}$. Mirror planes in $\{\mathbf{0}\}$ is ${}^\mathbf{0}\pi_\mathrm{Mj}= {}_{0}^{\mathbf{W}}\mathbf{T}^\mathsf{T}\pi_\mathrm{Mj}$ which contains two angles of rotation and one out-of-plane translation that can be mapped to Hall sensor readings in the next step. 

\subsection{Hall Sensor Calibration}
In Hall sensor data processing, we first linearly interpolate actual and background readings $\mathrm{B}_a$ and $\mathrm{B}_b$ as $\mathrm{B}_\mathrm{A}(t)$ and $\mathrm{B}_\mathrm{B}(t)$ to allow time offset estimation in model calibration~\cite{Rehder'16}. Then subtract the background signal from the actual signal to obtain the foreground signal as $\mathrm{B}(t) = \mathrm{B}_\mathrm{A}(t) - \mathrm{B}_\mathrm{B}(t)$ \cite{Son'16}.
With background interference removed, let us model and establish the mapping between the magnetic field readings and mirror poses. The experiment setup that generates these readings will be explained in the next section.

\begin{figure}[ht]
    \centering
    \subfigure[]{\includegraphics[width=0.9in, viewport=0 100 115 182, clip]{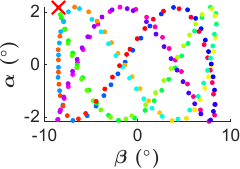} \label{fig:measurements_a}} \setcounter{subfigure}{2}
    \subfigure[]{\includegraphics[width=0.9in, viewport=130 100 245 182, clip]{image/PerceivedPatterns.pdf} \label{fig:measurements_c}} \setcounter{subfigure}{4}
    \subfigure[]{\includegraphics[width=0.9in, viewport=255 100 380 182, clip]{image/PerceivedPatterns.pdf} \label{fig:measurements_e}} \\ \setcounter{subfigure}{1}
    \vspace*{-.1in}
    \subfigure[]{\includegraphics[width=0.9in, viewport=2 0 115 90, clip]{image/PerceivedPatterns.pdf} \label{fig:measurements_b}} \setcounter{subfigure}{3}
    \subfigure[]{\includegraphics[width=0.9in, viewport=135 0 245 90, clip]{image/PerceivedPatterns.pdf} \label{fig:measurements_d}} \setcounter{subfigure}{5}
    \subfigure[]{\includegraphics[width=0.9in, viewport=260 0 380 90, clip]{image/PerceivedPatterns.pdf} \label{fig:measurements_f}}
    \caption{Angle measurements and Hall sensor readings in signal space (a and b) and temporal space (c to f). (a), (c), and (e) are mirror poses while (b), (d), and (f) are magnetic field readings with background removed. Time is coded in color. Red `X' shows one corresponding point across the two signals in different domains.} \label{fig:patterns}
\vspace*{-.1in}
\end{figure}
The angle measurements and Hall sensor readings are shown in \autoref{fig:patterns}. There is a clear correlation between the angle measurements and the Hall sensor readings. Because the sensing magnet is mounted on the back of mirror plate, its motion direction is always opposite to the mirror movement. This reversed motion is reflected in Figs.~\ref{fig:measurements_a} and \ref{fig:measurements_b}, a point (marked with red `X') in the top left corner of (a) corresponds to the bottom right corner of (b). Translating the correspondence in signal space to temporal space show the time offset between the two signals, the time offset can be observed by comparing Figs. \ref{fig:measurements_c} and \ref{fig:measurements_e} to Figs. \ref{fig:measurements_d} and \ref{fig:measurements_f}.

Based on the a near linear relationship and periodicity of the angle measurements and Hall sensor readings, we compare a linear model to a sine wave model in the mapping between the two types of signal.

A linear model maps a linear combination of the Hall sensor readings to the mirror plane as
\begin{equation}\label{eq:linear}
f_\mathrm{L}(\mathrm{A},\mathrm{B}_{j}) = \mathrm{A} [\mathrm{B}_{j}^{\mathsf{T}} \  1]^{\mathsf{T}}
\end{equation}
here $\mathrm{A}$ is a 3-by-4 matrix of model parameters, and $\mathrm{B}_{j}=\mathrm{B}(t_j)$.

Similarly, a sine wave model can be modeled as
\begin{equation}\label{eq:sin}
f_\mathrm{S}(\mathrm{A},\mathrm{B}_{j}) = \mathrm{A} [\sin(\Phi(\mathrm{B}_{j}))^{\mathsf{T}} \ 1]^{\mathsf{T}}
\end{equation}
here $\mathrm{A}$ is also a 3-by-4 matrix of model parameters. Phase mapping function $\Phi(\mathrm{B}_{j}) = 2\pi \mathbf{f}\mathrm{B}^{-1}(\mathrm{B}_{j})$, where $\mathbf{f}$ is a 3-by-1 vector of the foreground Hall effect sensor signal frequencies obtained from the data, $\mathrm{B}^{-1}$ is the inverse function of the Hall sensor reading interpolation that maps the readings back to time. We can now simultaneously estimate the time offset $\delta t$ that associates the two data sequences and the parameters of each model $f \in \{f_\mathrm{L},f_\mathrm{S}\}$ from
\begin{equation}\label{eq:hall_cali}
\min_{\mathrm{A}, \delta t} \sum_j \|f(\mathrm{A},\mathrm{B}(t_j+\delta t)) - {}^\mathbf{0}\tilde{\pi}_\mathrm{Mj} \|_2.
\end{equation}

\section{Experiments} \label{sec:experiment}

\subsection{Experiment Setup}
\begin{figure}[!htbp]
    \centering
    \includegraphics[width=2.5 in]{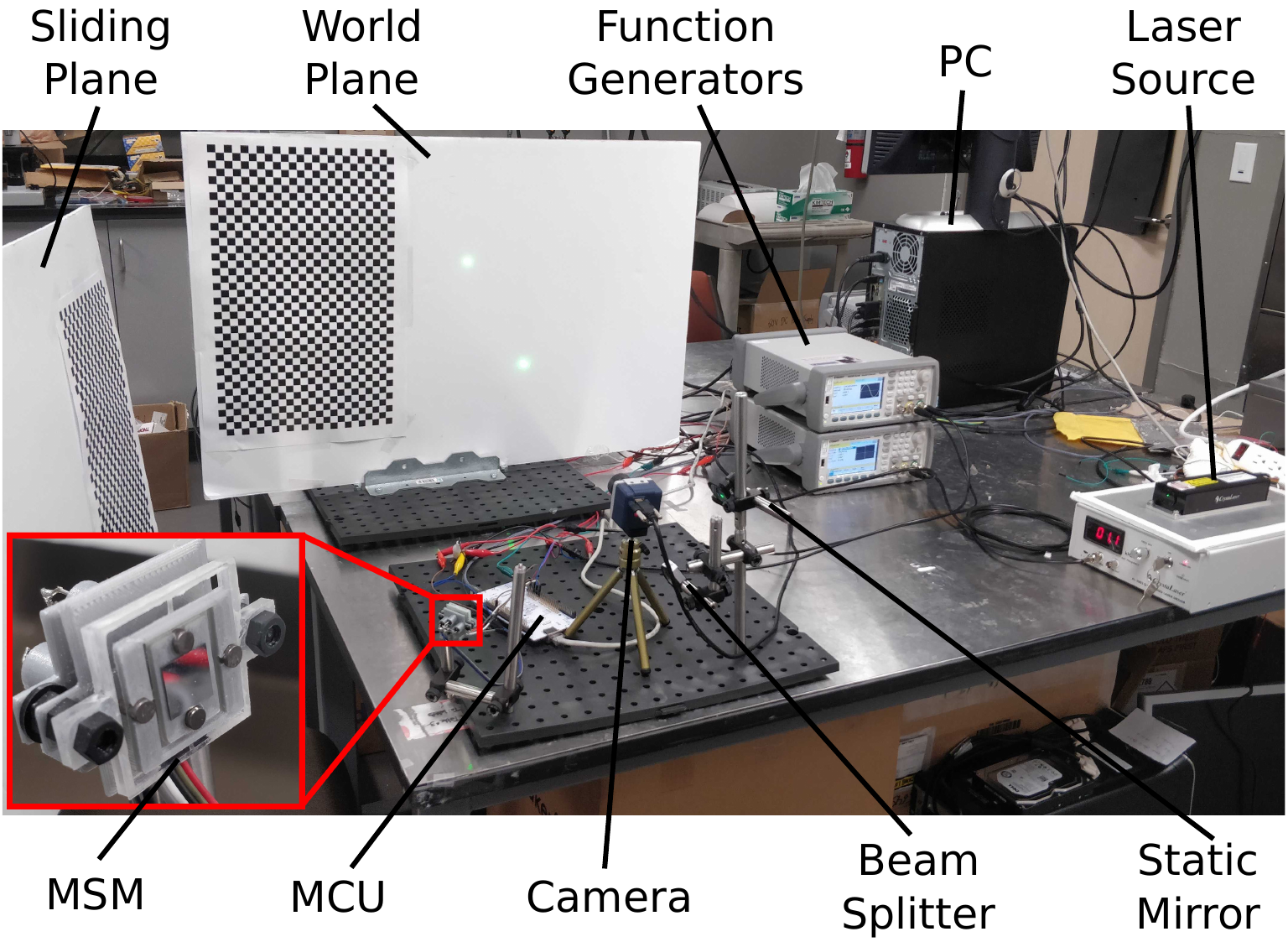}
    \caption{Photo of the experiment setup.} \label{fig:setup}
\end{figure}

The experiment setup is shown in \autoref{fig:setup}. Checkerboard patterns have a cell size of 10.0mm$\times$10.0mm. We employ two function generators (Keysight 33520B) to output driving and triggering signals. 
The pulse width of our laser source (Crystalaser QL532-1W0) is 15 ns. The sampling rate of our triaxial Hall effect sensor (Melexis MLX90393) is 1 kHz, and the MCU (STM NUCLEO-F439ZI) produces an interrupt signal when it receives a Hall sensor reading. We employ an industry-grade 10 mega-pixel CMOS camera (DS-CFMT1000-H) to capture images, and the camera intrinsic parameters have been calibrated using OpenCV. 

\subsection{Experimental Results}

\subsubsection{Mirror Pose Estimation Result} \label{Mirror_Est_Result}

We collected six datasets to validate the mirror pose estimation method in Sec.~\ref{ssc:MirrorPoseEst}. The six datasets are combinations of two different MSM scanning patterns and three laser beam setups with different incident angles. We compare the estimation results from our proposed 3 DoF mirror estimation method and a baseline method employed in \cite{Baier'22} that assumes pure MSM rotational motion and precise alignment between the MSM center of rotation and the incident laser beams.

We generate three laser beams using the beam splitter and a static mirror. We employ two beams to estimate mirror poses while validating the estimation with the third beam. The choice of the two beams causes different angles between the two incident beams and leads to different spanning angles $\theta$ between the light path plane normals (shown in Fig.~\ref{fig:WorkingPrinciple-normal}). $\theta$ changes during MSM scanning and as it approaches $90^{\circ}$ it reduces the uncertainty of mirror pose estimation; therefore, we use it as a control variable to validate our results, a larger average spanning angle $\bar{\theta}$ should lead to a smaller prediction error.

Because the ground truth measurements of the 3 DoF mirror poses are not available, we validate the estimated mirror poses by comparing the projections of the reflected beam on the world plane predicted by the estimations with the actual observations. The difference between the predictions and observations is calculated using (\ref{eq:MLE-PredictionError}) and converted from pixel space to angular space. We use the root-mean-squared angular errors $\bar{\delta}$ to measure the accuracy of our calibration.

\begin{table}[!htbp]
	\centering
	\caption{Mirror Pose Estimation Result}\label{tab:MirrorEstResult}
  \resizebox{\hsize}{!}{
	  \begin{tabular}{c c c c}
		  \toprule
		  \multirow{2}{*}{Dataset} & Avg. Spanning Angle & Baseline Model Error & Proposed Model Error \\
		        & $\bar{\theta} ({}^\circ)$ & $\bar{\delta}_b\pm\sigma_b$ ($^\circ$) & $\bar{\delta}_p\pm\sigma_p$ ($^\circ$)    \\
		  \midrule
		  Pattern-A-1 & 18.45 & $0.045\pm 0.023$ & $0.042\pm 0.026$ \\
	   	Pattern-A-2 & 27.67 & $0.121\pm 0.076$ & $0.031\pm 0.020$ \\
		  Pattern-A-3 & 46.13 & $0.036\pm 0.018$ & $0.022\pm 0.012$ \\
		  Pattern-B-4 & 18.34 & $0.106\pm 0.071$ & $0.032 \pm 0.017$\\
		  Pattern-B-5 & 27.15 & $0.133\pm 0.093$ & $0.030 \pm 0.008$\\
		  Pattern-B-6 & 45.49 & $0.061\pm 0.034$ & $0.020 \pm 0.011$\\
		  \bottomrule
	  \end{tabular}
  }
\end{table}

As shown in Tab.~\ref{tab:MirrorEstResult}, the maximum prediction error of the proposed model is 0.042$^{\circ}$ in the six datasets, and the minimum is 0.020$^{\circ}$. The proposed 3-DoF mirror estimation model performs better than the baseline model in all six datasets.

The proposed model also shows a consistent error level when choosing the same spanning-angle laser between the two incident beams used in calibration. As shown in Tab.~\ref{tab:MirrorEstResult}, for the Pattern-A-3 and Pattern-B-6 datasets, when the averaged spanning angles $\bar{\theta}$ between normals are greater than $45^\circ$, the prediction errors are the smallest between the datasets. This is expected because a larger angle between the light-path plane normals leads to a smaller uncertainty range of the mirror plane estimations; hence the higher accuracy improves. The detailed error analysis results are included on page 4 of the multimedia attachment file. 

\subsubsection{Hall Sensor Calibration Result}
We have collected a dataset with 195 mirror pose measurements and 130k actual Hall sensor readings and 65k background Hall sensor readings. The dataset is randomly divided into a training set and a testing set with a ratio of 4:1. Both the linear model and the sine wave model have been fitted into the training set by solving \eqref{eq:hall_cali} for the two candidate models in \eqref{eq:linear} and \eqref{eq:sin}. Each estimated model is then used to predict the mirror planes with Hall sensor readings in the testing set, and the errors between the predictions and measurements are used for comparison. We repeat this process 50 times and compare the root mean squared error of each model. To maintain independence of the training and testing process, Hall effect sensor readings are interpolated separately on the training and testing sets. In the dataset, the range of mirror scanning angles is $4.37^\circ$ and $17.17^\circ$ for the fast and slow axes, respectively. The range of out-of-plane translation is $1.04$ mm.

The root-mean-squared test errors of the two models on the 50 random trails are shown in Tab.~\ref{tab:CalibResult}. The sine wave model performs better than the linear model, which is not surprising because the sine wave model captures the inherent periodicity property better than the linear model. It is also expected to be more robust to the baseline shift caused by external magnetic interference.

\begin{table}[!htbp]
	\centering
	\caption{Hall Sensor Calibration Result}\label{tab:CalibResult}
	\begin{tabular}{c c c c}
		\toprule
		Model      & $\bar{\delta}_\alpha\pm\sigma_\alpha$ ($^\circ$) & $\bar{\delta}_\beta\pm\sigma_\beta$ ($^\circ$) & $\bar{\delta}_d\pm\sigma_d$ (mm) \\
		\midrule
	    Linear     & $0.101\pm 0.053$ & $0.082\pm 0.052$ & $0.11\pm 0.08$ \\
        \midrule
		Sine Wave  & $ 0.083\pm 0.040$ & $0.069\pm 0.040$ & $0.10\pm 0.08$ \\
		\bottomrule
	\end{tabular}
\end{table}

\vspace{-0.5cm}
\section{Conclusions and Future Work}

We reported on our design of a calibration rig and algorithms for MSM with triaxial Hall sensors. To reduce cost and address the unique challenges brought by MSMs, we employed a 2-laser beam approach assisted by two checkerboards. We extracted laser dot patterns and modeled their reflection property to propose an indirect mirror pose estimation method. We also proposed a self-synchronizing optimization approach that exploits the signal periodicity to map mirror poses to Hall sensor readings. We constructed the calibration rig and implemented algorithms. Our experimental results validated our design with satisfactory results. In the future, we will further explore optimal calibration setup (e.g. incident beam number and spanning angles) and calibrate MSMs with different sensory feedback methods. New results will be reported in future publications. 

\vspace{-0.8cm}
\section*{}
{\small
\section*{Acknowledgment}
We are grateful to Y. Ou, A. Kingery, S. Xie, F. Guo and C. Qian for their inputs and contributions.

\bibliographystyle{fmt/IEEEtran}
\bibliography{bib/di}
}

\begin{appendices}
\section{Derivations for Section V}

In the following section, we present the detailed derivation of $f_P$, $f_L$ and $f_R$ appearing in (\ref{eq:Proj-Cost}), (\ref{eq:Line-Cost}) and (\ref{eq:Reflection-Cost}).

For the PnP constraint $f_{P}(\mathrm{\tilde{T}},\mathrm{X})$, recall that it is from the projection relationship in (\ref{eq:PnPConstrain}), which projects a 3D point $\mathrm{X}$ to 2D based on camera intrinsic matrix $\mathbf{K}$ and camera pose defined by $\mathrm{\tilde{T}}$. Its input $\mathrm{X} = [\mathrm{\tilde{X}}^\mathsf{T} \textrm{  }  1]^\mathsf{T}$ is the homogeneous counterpart of 3D point $\mathrm{\tilde{X}} \in \mathbb{R}^3 $, and $\mathrm{\tilde{T}} = [\mathbf{w}^\mathsf{T} \ \mathbf{t}^\mathsf{T}]^\mathsf{T}$ is the minimum parameterization of world-to-camera transformation matrix $\mathbf{T} = \begin{bmatrix}\mathbf{R} & \mathbf{t} \\ \mathbf{0} & 1\end{bmatrix}$. $\mathbf{w}\in\mathbb{R}^3$ is the axis-angle parameterization of rotation matrix $\mathbf{R}$, and according to Rodrigues' formula $\mathbf{R} = \mathbf{I} + \sin(\|\mathbf{w}\|_2)[\mathbf{w}]_\times+(1-\cos(\|\mathbf{w}\|_2))[\mathbf{w}]^2_\times$, where $\|\cdot\|_2$ is the $L_2$ norm and $[\cdot]^2_\times$ is the skew-symmtric matrix. Therefore, $f_{P}(\mathrm{\tilde{T}},\mathrm{X})$ can be derived as 
\begin{align} \label{eq:f_P}
\begin{split} 
    f_{P}(\mathrm{\tilde{T}},\mathrm{X}) &= \lambda \mathbf{K}[\mathbf{R} \textrm{  } \mathbf{t}]\mathrm{X} \\
&= \lambda \mathbf{K}(\mathbf{R}\mathrm{\tilde{X}}+\mathbf{t}) \\
&= \lambda \begin{bmatrix} f_x & & c_x \\  & f_y & c_y \\ & & 1\end{bmatrix} \left ( \begin{bmatrix} \mathbf{r}^\mathsf{T}_1  \\ \mathbf{r}^\mathsf{T}_2 \\ \mathbf{r}^\mathsf{T}_3\end{bmatrix}\mathrm{\tilde{X}}+\begin{bmatrix} t_x  \\ t_y \\ t_z\end{bmatrix}\right ) \\
&= \frac{1}{\mathbf{r}^\mathsf{T}_3\mathrm{\tilde{X}}+t_z} \begin{bmatrix} f_x(\mathbf{r}^\mathsf{T}_1\mathrm{\tilde{X}}+t_x)+c_x(\mathbf{r}^\mathsf{T}_3\mathrm{\tilde{X}}+t_z)  \\ f_y(\mathbf{r}^\mathsf{T}_2\mathrm{\tilde{X}}+t_y)+c_y(\mathbf{r}^\mathsf{T}_3\mathrm{\tilde{X}}+t_z)  \\ \mathbf{r}^\mathsf{T}_3\mathrm{\tilde{X}}+t_z\end{bmatrix}
\end{split}
\end{align}
Here $\mathbf{r}^\mathsf{T}_i$ is the i-th row of $\mathbf{R}$. 

For the line constraint $f_{L}({}_{\textit{S}_l}^{\textit{C}_1}\mathrm{\tilde{T}},\mathrm{\tilde{L}}_i,{}_{\textit{W}}^{\textit{S}_l}\mathrm{\tilde{T}})$, recall that it is from the projection relationship in (\ref{eq:PnPConstrain}) and the 3D point-on-line relationship in (\ref{eq:LineConstrain}). It is the 2D projection of the intersecting 3D point of i-th laser beam $\textrm{L}_i$ and l-th sliding plane $\pi_l$ in camera $\textit{C}_1$, therefore, we will derive $\textrm{L}_i$ and $\pi_l$ first. The inputs of $f_{L}$, ${}_{\textit{S}_l}^{\textit{C}_1}\mathrm{\tilde{T}}$ and ${}_{\textit{W}}^{\textit{S}_l}\mathrm{\tilde{T}}$ correspond to the minimum parameterization of the l-th sliding-plane-to-camera and world-to-sliding-plane transformation matrix, respectively. And $\mathrm{\tilde{L}}_i = [\mathbf{w}_{Li}^\mathsf{T} \textrm{  }  m_i]^\mathsf{T}$ is the minimum parameterization of the i-th laser beam $\mathbf{L}_i = [[\mathbf{v}_i]_\times \textrm{  }  \mathbf{m}_i]$. By the parameterization defined in \cite{Bartoli'05}, $\mathbf{R}_{Li} =  \begin{bmatrix}\mathbf{v}_i & \frac{\mathbf{m}_i}{\|\mathbf{m}_i\|_2} & \mathbf{v}_i \times \frac{\mathbf{m}_i}{\|\mathbf{m}_i\|_2} \end{bmatrix}$, and it can be recover from $\mathbf{w}_{Li}$ using the Rodrigues' formula. The l-th sliding plane $\pi_l$ can be obtained from ${}_{\textit{W}}^{\textit{S}_l}\mathrm{T}$ and world plane $\pi_W = [0 \ 0 \ 1 \ 0]^\mathsf{T}$ as $\pi_l = {}_{\textit{W}}^{\textit{S}_l}\mathrm{T}^\mathsf{T}\pi_W = [{}_{\textit{W}}^{\textit{S}_l}\mathbf{r}_3^\mathsf{T} \ {}_{\textit{W}}^{\textit{S}_l}t_z]^\mathsf{T}$. The intersecting 3D point $\mathrm{X}_{i,l}$ of i-th laser beam $\textrm{L}_i$ and l-th sliding plane $\pi_l$ satisfies $\begin{bmatrix}\textrm{L}_i \\ \pi_l^\mathsf{T}\end{bmatrix}\mathrm{X}_{i,l} = \mathbf{0}$, therefore
\begin{equation} \label{eq:X_il}
\begin{split} 
    \mathrm{\tilde{X}}_{i,l} = - \begin{bmatrix}[\mathbf{v}_i]_\times \\ {}_{\textit{W}}^{\textit{S}_l}\mathbf{r}_3^\mathsf{T}\end{bmatrix}^+\begin{bmatrix}\mathbf{m}_i \\ {}_{\textit{W}}^{\textit{S}_l}t_z\end{bmatrix}
\end{split}
\end{equation}
where $[\cdot]^+$ is the pseudo-inverse. With the 3D intersecting point $\mathrm{X}_{i,l}$ of i-th laser beam $\textrm{L}_i$ and l-th sliding plane $\pi_l$, $f_{L}({}_{\textit{S}_l}^{\textit{C}_1}\mathrm{\tilde{T}},\mathrm{\tilde{L}}_i,{}_{\textit{W}}^{\textit{S}_l}\mathrm{\tilde{T}})$ can be derived as 
\begin{align} \label{eq:f_L}
\begin{split} 
    f_{L}({}_{\textit{S}_l}^{\textit{C}_1}\mathrm{\tilde{T}},\mathrm{\tilde{L}}_i,{}_{\textit{W}}^{\textit{S}_l}\mathrm{\tilde{T}}) = f_{P}({}_{\textit{W}}^{\textit{C}_1}\mathrm{\tilde{T}},\mathrm{X}_{i,l})
\end{split}
\end{align}
Here ${}_{\textit{W}}^{\textit{C}_1}\mathrm{\tilde{T}}$ is obtained from ${}_{\textit{S}_l}^{\textit{C}_1}\mathrm{T}{}_{\textit{W}}^{\textit{S}_l}\mathrm{T}$.

For the reflection constraint $f_{R}({}_{\textit{W}}^{\textit{C}_2}\mathrm{\tilde{T}},\mathrm{\tilde{L}}_i,\tilde{\pi}_\mathrm{Mj})$, recall that it is obtained from the projection relationship in (\ref{eq:PnPConstrain}) and the 3D reflected-point-on-line relationship in (\ref{eq:ReflectionConstrain}). It is the 2D projection of the intersecting 3D point of j-th mirror pose reflected i-th laser beam $\textrm{L}_i\mathbf{H}_j$ and world plane $\pi_W$ in camera $\textit{C}_2$. The deravation of i-th reflected laser beam $\textrm{L}_i\mathbf{H}_j$ is similar to its counterpart in (\ref{eq:f_L}), $\textrm{L}_i = [[\mathbf{v}_i]_\times \textrm{  }  \mathbf{m}_i]$ is first recovered from its minimum parameterization $\mathrm{\tilde{L}}_i$. Then the j-th mirror plane $\pi_\mathrm{Mj}$ is obtained from $\tilde{\pi}_\mathrm{Mj}$ as $\pi_\mathrm{Mj} = [\mathbf{n}^\mathsf{T}_\mathrm{Mj} \ d_\mathrm{Mj}]^\mathsf{T} =\frac{1}{\left \Arrowvert \tilde{\pi}_\mathrm{Mj}\sin\frac{\|\tilde{\pi}_\mathrm{Mj}\|_2}{2}\right \Arrowvert_2}\begin{bmatrix}\tilde{\pi}^\mathsf{T}_\mathrm{Mj}\sin\frac{\|\tilde{\pi}_\mathrm{Mj}\|_2}{2} & \cos\frac{\|\tilde{\pi}_\mathrm{Mj}\|_2}{2} \end{bmatrix}$, and the corresponding reflection matrix $\mathbf{H}_j = \begin{bmatrix} \mathbf{I}-2\mathbf{n}_\mathrm{Mj}\mathbf{n}_\mathrm{Mj}^\mathsf{T} & -2d_\mathrm{Mj}\mathbf{n}_\mathrm{Mj} \\ \mathbf{0} & 1\end{bmatrix}$. With the j-th mirror pose reflected i-th laser beam $\textrm{L}_i\mathbf{H}_j$ and world plane $\pi_W$, their intersecting 3D point $\mathrm{X}_{i,j}$ satisfies $\begin{bmatrix}\textrm{L}_i\mathbf{H}_j \\ \pi_W^\mathsf{T}\end{bmatrix}\mathrm{X}_{i,j} = \mathbf{0}$, therefore
\begin{equation} \label{eq:X_ij}
\begin{split} 
    \mathrm{\tilde{X}}_{i,j} = - \begin{bmatrix}[\mathbf{v}_i]_\times - 2[\mathbf{v}_i]_\times\mathbf{n}_\mathrm{Mj}\mathbf{n}_\mathrm{Mj}^\mathsf{T} \\ \mathbf{n}_W^\mathsf{T}\end{bmatrix}^+(2d_\mathrm{Mj}[\mathbf{v}_i]_\times\mathbf{n}_\mathrm{Mj}-\mathbf{m}_i)
\end{split}
\end{equation}
where $\mathbf{n}_W^\mathsf{T} = [0 \ 0 \ 1]$ by definition. With the 3D intersecting point $\mathrm{X}_{i,j}$, $f_{R}({}_{\textit{W}}^{\textit{C}_2}\mathrm{\tilde{T}},\mathrm{\tilde{L}}_i,\tilde{\pi}_\mathrm{Mj})$ can be derived as 
\begin{align} \label{eq:f_L}
\begin{split} 
    f_{R}({}_{\textit{W}}^{\textit{C}_2}\mathrm{\tilde{T}},\mathrm{\tilde{L}}_i,\tilde{\pi}_\mathrm{Mj}) = f_{P}({}_{\textit{W}}^{\textit{C}_2}\mathrm{\tilde{T}},\mathrm{X}_{i,j})
\end{split}
\end{align}

\section{Supplementary Figures}
\begin{figure}[ht]
    \centering
    \subfigure[]{\includegraphics[width=0.9in, viewport=0 80 115 154, clip]{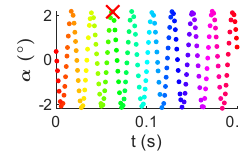} \label{fig:measurements_a1}}
    \subfigure[]{\includegraphics[width=0.9in, viewport=130 80 245 154, clip]{image/PerceivedPatterns-1.pdf} \label{fig:measurements_b1}}
    \subfigure[]{\includegraphics[width=0.9in, viewport=255 80 380 154, clip]{image/PerceivedPatterns-1.pdf} \label{fig:measurements_c1}}\\
    \vspace*{-.1in}
    \subfigure[]{\includegraphics[width=0.9in, viewport=0 0 115 80, clip]{image/PerceivedPatterns-1.pdf} \label{fig:measurements_d1}}
    \subfigure[]{\includegraphics[width=0.9in, viewport=135 0 245 80, clip]{image/PerceivedPatterns-1.pdf} \label{fig:measurements_e1}} 
    \subfigure[]{\includegraphics[width=0.9in, viewport=248 0 380 80, clip]{image/PerceivedPatterns-1.pdf} \label{fig:measurements_f1}}

    \caption{Mirror pose measurements and Hall sensor readings in temporal space. (a), (b), and (c) are mirror poses while (d), (e), and (f) are magnetic field readings with background removed. Time is coded in color. Red `X' shows one corresponding point across the two signals in different domains.} \label{fig:patterns}
\vspace*{-.1in}
\end{figure}
\end{appendices}

\end{document}